**Title:** Automatic sampling and training method for wood-leaf classification based on tree terrestrial point cloud

**Author:** Zichu Liu[1], Qing Zhang[1*], Pei Wang[1*], Yaxin Li[1], Jingqian Sun[1]

**Address:** [1]College of Science, Beijing Forestry University, No.35 Qinghua East Road, Haidian District, Beijing 100083, China

**Corresponding author:** Qing Zhang[1*], Pei Wang[1*]

**E-mail:** Zichu Liu: lzc0330@bjfu.edu.cn

Qing Zhang: zhangq@bjfu.edu.cn

Pei Wang: wangpei@bjfu.edu.cn

Yaxin Li: liyaxin_2018@bjfu.edu.cn

Jingqian Sun: sunjq_2019@bjfu.edu.cn

[*]**For correspondence. E-mail:**
zhangq@bjfu.edu.cn
wangpei@bjfu.edu.cn

**Abstract**

Terrestrial laser scanning technology provides an efficient and accuracy solution for acquiring three-dimensional information of plants. The leaf-wood classification of plant point cloud data is a fundamental step for some forestry and biological research. An automatic sampling and training method for classification was proposed based on tree point cloud data. The plane fitting method was used for selecting leaf sample points and wood sample points automatically, then two local features were calculated for training and classification by using support vector machine (SVM) algorithm. The point cloud data of ten trees were tested by using the proposed method and a manual selection method. The average correct classification rate and kappa coefficient are 0.9305 and 0.7904, respectively. The results show that the proposed method had better efficiency and accuracy comparing to the manual selection method.

## 1 Introduction

Forests have direct impacts on global environment and climate change (Bonan, 2008). Woods and leaves are two important components of tree biomass. The information of the former can be used to estimate the volume amount of a tree, and the later can be used for the inference of leaf area index (Keramatlou et al. 2015). Effective and accurate classification of woods and leaves make contributions to estimating the biomass information of trees and analyzing other parameters, which can also help for estimating the impact of forests on the environment and climate.

The traditional forest investigation methods are usually manual measurement, which is inefficient and time consuming (Gibbs et al. 2018). TLS is an accurate and reliable measurement method in forestry inventories because of its ability of acquiring the high-density and high-precision point cloud data (Yan et al., 2015). The point cloud data obtained by using TLS was used to estimate tree characteristics such as tree structure information (Liang et al., 2016; Dassot et al., 2011), leaf area index (LAI) (Zheng et al., 2013; Kim et al., 2009; Zhao and Popescu, 2009; Lange et al., 2009; Hosoi et al., 2010), leaf area density (Béland et al., 2014) and structure information of forest (Lovell et al., 2003; Newnham et al., 2015). TLS was also used for tree reconstruction (Chen et al., 2010; Garrido, et al. 2015), forest inventory parameter analysis (Maas et al., 2008) and forest biomass estimation (Popescu, 2007).

The classification of leaves and wood is not only an essential prerequisite of most of the above research, but also a potential research direction. The traditional methods used in research about leaf and wood measuring and classification are usually destructive sampling, which is harmful to trees. In recent years, some classification research has been done on the point cloud data of plants. Paulus et al. (2013) introduced an adapted surface feature based method to classify grapevine point clouds to leaf and stem organs. Tao et al. (2015) classified the leaf and stem points of trees by using the shortest path method and the axis transformation method based on the spatial coordinates of the point cloud data. Zheng et al. (2016) proposed a method based on local geometric features by identifying the differences between directional gap fraction and angular gap fraction to classify forest point cloud data. Ferrara et al. (2018) proposed a leaf-wood separation method based on spatial geometric information, the point cloud data were partitioned into voxels and clustered by using DBSCAN algorithm. Yun et al. (2016) constructed sphere neighborhoods to extract multiple features, then classified trees point cloud data to leaves and wood by using support vector machine (SVM) algorithm. Vandapel et al. (2004) used the Bayes classifier to separate the point cloud data into three categories (surface, linear structures and scatter) through the spatial distribution of local neighborhood. Zhu et al. (2018) classified mixed natural forest point cloud data to leaves and wood by using random forest (RF) algorithm. Jin et al. (2019) proposed a median normalized-vector growth algorithm, and classified leaves and stems of 30 maize samples with four steps. Xiang et al. (2019) used a skeletonization method to classify the stem and leaves of sorghum plants.

Many of the above-mentioned research about leaf-wood classification require traditional and manual sampling and separation, which cause much time cost and labor cost, and also irreversible destruction to plant samples. Meanwhile, the classification results are limited by subjective factors when sampling manually. Therefore, some authors set about studying automatic classification method to improve the precision and efficiency.

Ma et al. (2016) developed an automatic point classification method based on geometric information contained in point cloud data to separate photosynthetic and non-photosynthetic canopies of forest. Liu et al. (2020) proposed an automated method for leaf-stem classification of potted plant point cloud data, the 3D convex hull algorithm was applied for automated selection of leaf sample points and the projection densities were used for selecting stem sample points automatically. Then the SVM algorithm was used for classification of the leaf points and stem points of three potted plants.

Nevertheless, the related research either need partial manual operations, or have limitations on some specific species or shapes of plants.

In this paper, we proposed an automatic method for classifying tree point cloud data. This method consists of sampling module and classification module. Local geometric features were used for constructing training sets automatically and SVM algorithm was used to classify point cloud data into leaf points and wood points.

2 Materials

The study area is located in Haidian Park, Beijing. Ten experimental trees were selected and scanned by using VZ-400 TLS (*RIEGL* Laser Measurement Systems GmbH, 3580 Horn, Austria). This device can scan high-precision three-dimensional data effectively, the specific information of this scanner was listed in Table 1.

| 3D Terrestrial Laser Scanner | *REIGL* VZ-400 |
|---|---|
| Largest Measurement Range | 600 $m$ |
| Highest Measurement Rate | 122,000 measurements / second |
| Measurement Accuracy | 2 $mm$ |
| Laser Emission Frequency | 300,000 points / second |
| The vertical field of view | 100° |
| The horizontal field of view | 360° |
| Connection | LAN/WLAN, wireless data transmission |
| Operational control | Desktop,PDA or Laptop |

Table 1. Information of *REIGL* VZ-400 scanner.

Each tree was scanned in single-site with an angular step-width of 0.02 degree in both vertical and horizontal directions. And then, as shown in Fig. 1, the tree point clouds were extracted separately.

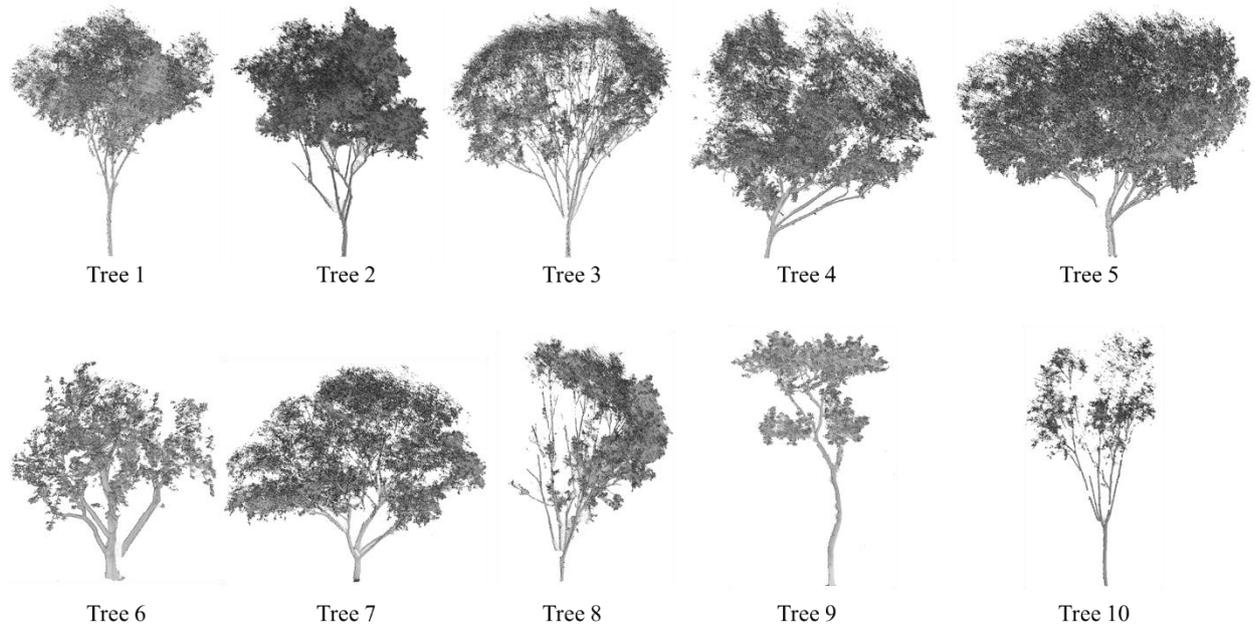

Fig. 1. Point cloud data of ten trees.

3 Method

The steps of this work are as follows (as shown in Fig.2): (1) Ten trees were scanned and the tree point clouds were extracted. (2) Some local features were proposed and calculated for classification. (3) The standard classification results were constructed. (4) Tree point clouds were classified into leaf points and wood points by using the proposed method. (5) Finally, the manually classified results were generated for comparison and discussion.

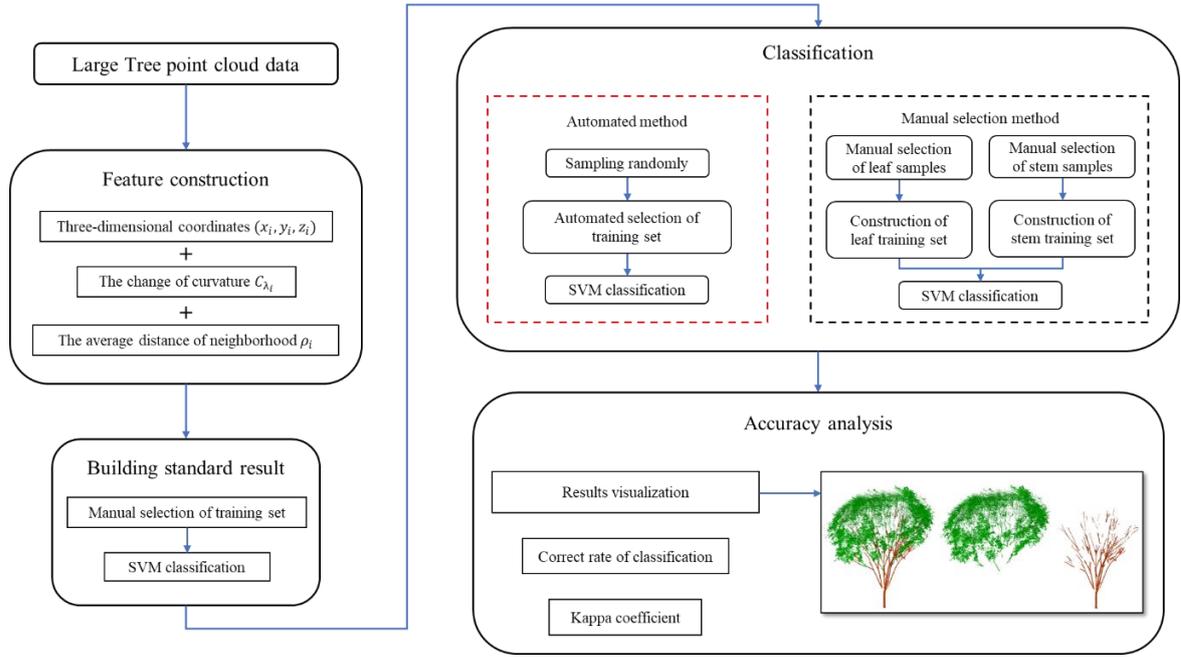

Fig. 2. Flowchart of experiment.

3.1 Feature extraction

Although the geometric and intensity information are both scanned in the experiment, after many attempts, the intensity information is not included in the method. Therefore, the change of curvature and density feature of local neighborhood were computed and used for better training and classification results.

The specific construction processes of features are as follows:

First, for a point $p_{i_0} = (x_i, y_i, z_i)$ in point cloud data ($i = 1, \cdots, N$, where $N$ denotes the points number of point cloud data), k-Nearest Neighbor (kNN) method was used for constructing the local neighborhoods.

Second, the covariance matrix was calculated based on the $p_{i_0}$ neighborhoods (Yun et al., 2016):

$$C_{p_i} = \frac{1}{k+1}\sum_{j=0}^{k}(p_{i_j} - \bar{p}_i)(p_{i_j} - \bar{p}_i)^T \tag{1}$$

where $\bar{p}_i = \frac{1}{k+1}\sum_{j=0}^{k} p_{i_j}$.

Then, the eigenvalues $\lambda_{i_1}, \lambda_{i_2}, \lambda_{i_3}$ ($\lambda_{i_1} > \lambda_{i_2} > \lambda_{i_3}$) of $C_{p_i}$ were computed and normalized:

$$e_{i_s} = \frac{\lambda_{i_s}}{\lambda_{i_1}+\lambda_{i_2}+\lambda_{i_3}}, s = 1,2,3. \quad (2)$$

Next, the change of curvature $C_{\lambda_i}$ can be computed as (Ni et al., 2016):

$$C_{\lambda_i} = \frac{e_{i_3}}{e_{i_1}+e_{i_2}+e_{i_3}}. \quad (3)$$

After that, the average distance between neighborhood points to $p_{i_0}$ was also calculated to indicate the density of k-nearest neighborhoods:

$$\rho_i = \frac{1}{k}\sum_{m=1}^{k} d_m \quad (4)$$

where $d_m$ indicates the distance between point $p_{i_m}$ to point $p_{i_0}$.

Finally, using above two features and the three-dimensional coordinates $(x, y, z)$, the features of each point were obtained as $(x_i, y_i, z_i, C_{\lambda_i}, \rho_i)$. Then they were used for training and classification by SVM classifier.

3.2 Construction of results for comparison

Because of the large number of trees point cloud data, it is difficult to directly set up a classification standard by classifying the point cloud data manually. In this paper, the classification criteria were established by using SVM algorithm.

1. First, ten thousand points were selected randomly for each tree.

2. Then, these points were marked into leaf points and wood points manually.

3. Next, points at leaves and trunks were regarded as training points with different labels.

4. Finally, the SVM algorithm was applied based on the above training point sets, and the classification results were used as standard results (as shown in Fig. 3).

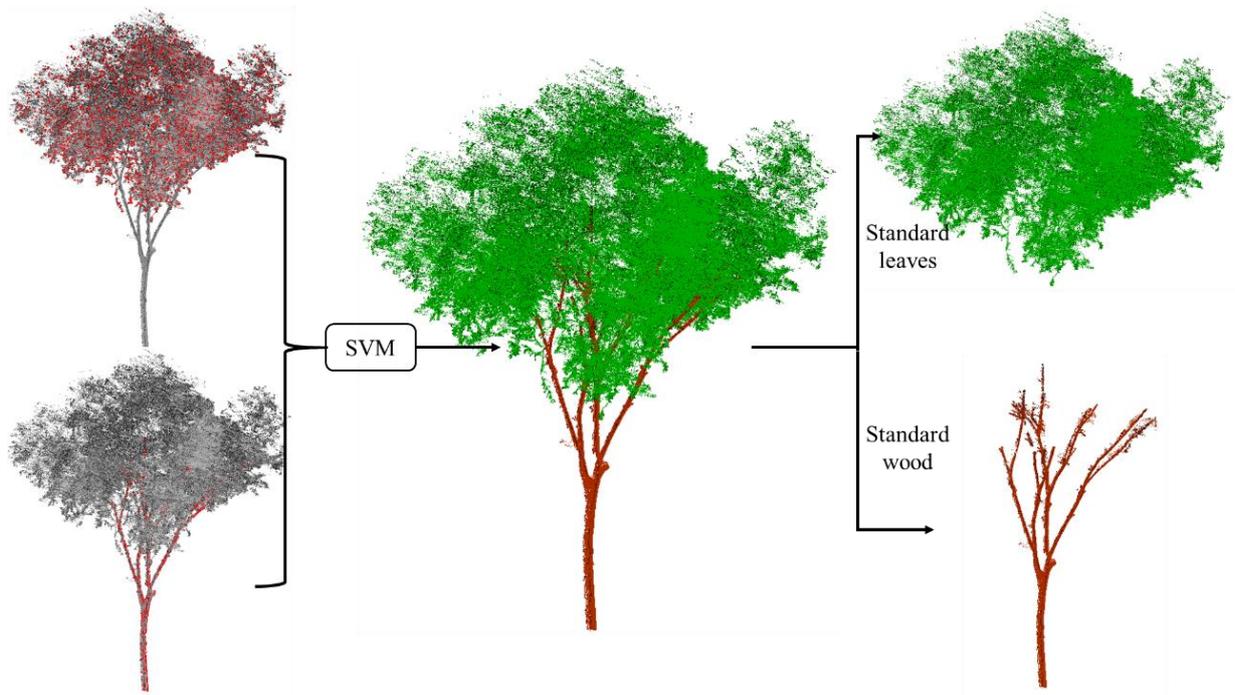

Fig. 3. Construction of standard results.

To discuss the proposed method and draw conclusion, a traditional sampling method was used for comparison. Liu et al. (2020) used a manual selection method for evaluating the effect of their method, the results showed that this method have good performances in accuracy, which is close to the standard results, hence it was used in this paper for comparison:

1. Twenty leaf points and twenty wood points with even distributions were sampled manually as seed points.

2. Spherical neighborhoods taken seed points as centers were constructed.

3. Then points inside the leaf spherical neighborhoods were regarded as the leaf training sets and those inside the wood spherical neighborhoods were regarded as wood training sets.

4. The training sets of leaf and wood were taken into SVM classifier for training and classification.

5. Finally, the classification results were used for comparison with the proposed method.

### 3.3 Sampling automatically and classification

Due to the morphological characteristics of the different organs of trees, we adopted the KNN searching method to construct neighborhoods, and plane fitting method to automatically select the training points of leaf and wood in the proposed method.

First, 2000 points were selected automatically, then k-nearest neighborhoods of each point were constructed. Then a plane was fitted based on the neighborhoods by using the least square method. The standard deviation of the distances from neighborhood points to the plane was computed.

Theoretically, the standard deviation of point located in leaves was much higher than it of point located in wood because the wood points are closer to a plane while the leaf points are more sparsely distributed. Therefore, the points with higher standard deviations were selected as leaf training points and the points with smaller standard deviations were regarded as wood training points.

For example, as shown in [Fig. 4](), point $a_l$ and point $a_s$ are located on leaves and wood, respectively. The $plane$ 1 was fitted based on the $a_l$ k-nearest neighborhood and the $plane$ 2 was fitted based on the $a_s$ k-nearest neighborhood, in which $k = 100$. As shown in [Fig. 4](), the neighborhood of leaf point $a_l$ have a more separated distribution around the $plane$ 1 but the neighborhood of wood point $a_s$ is more concentrated on the $plane$ 2.

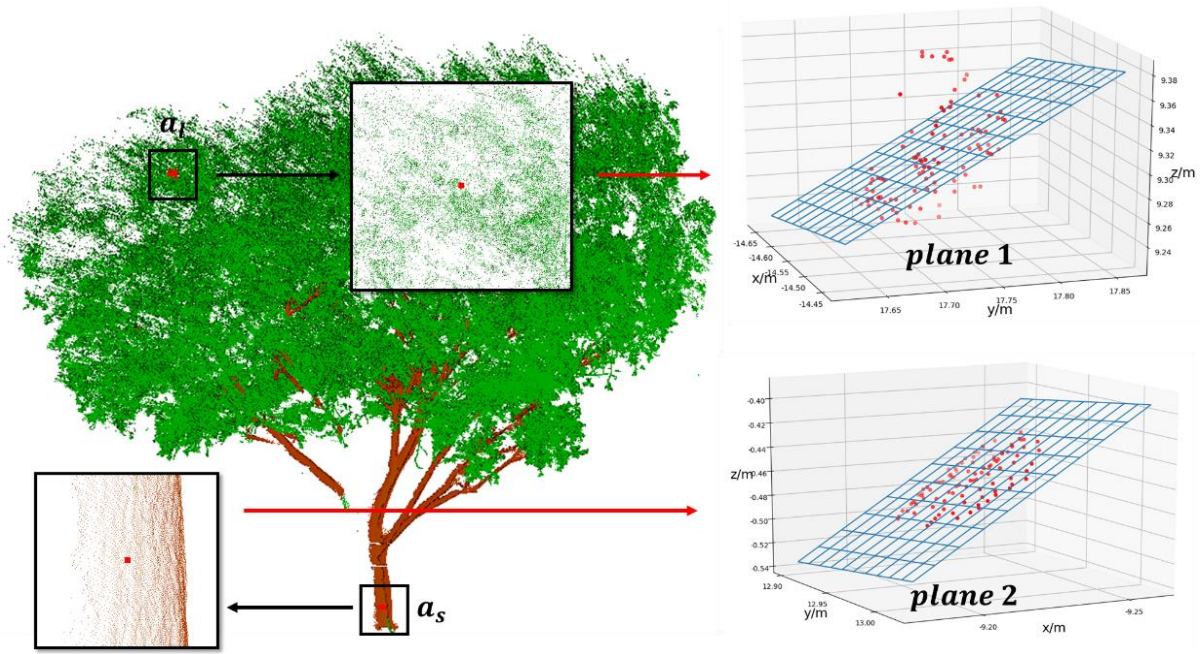

Fig. 4. Automatic sampling process.

Based on the leaf and wood training points selected automatically, the SVM classifier with radial basis function (RBF) kernel was used for classifying point cloud data of trees into leaf points and wood points. The SVM is a machine learning algorithm proposed by Vapnik (Vapnik, 1999). It can map the data to a space with higher dimensions, then classify the data by constructing hyperplanes.

The leaf training sets with features $(x_i, y_i, z_i, C_{\lambda_i}, \rho_i)$ were marked as class 1, and the wood training sets with these 5 features were marked as class 2. Then they were taken for training the classifier. Finally, the point cloud data were classified into leaf points and wood points based on these features by using this SVM classifier.

3.4 Accuracy evaluation

Accuracy of the proposed method was assessed by comparing the results of different methods. First, some indicators were calculated, which were the number of correctly classified leaf points $TP$, the number of correctly classified wood points $TN$, the number of mistakenly classified leaf points $FP$, and the number of mistakenly classified wood points $FN$, respectively.

Then, the correct classification rates $p_o$ can be computed:

$$p_o = \frac{TP+TN}{N} \tag{5}$$

where $N$ denotes the total points number of point cloud data. In addition, the kappa coefficient (Cohen, 1960) $kappa$ was also used because of its widely application in related research and the ability for summarizing classification results of imbalanced data:

$$kappa = \frac{p_o - p_e}{1 - p_e} \tag{6}$$

where $p_e = \frac{(TP+FP) \times (TP+TN) + (TN+FN) \times (FP+FN)}{N \times N}$.

4 Results

4.1 Sampling results

The sampling results were selected automatically by using the proposed method. After experimental tests, $k$ was set to 100 for KNN searching. Considering the different morphological characteristics of trees, the numbers of leaf training points and wood training points were confirmed. Tree 1 to Tree 7 have larger volumes of leaves than wood, so 1200 points of leaf training set and 800 points of wood training set were selected. Tree 8 and Tree 9 have leaves and wood with same volumes, 1000 points for each organ's training set were chosen. The leaves of Tree 10 occupy smaller space than the wood, which made 800 points and 1200 points were selected for leaf training set and wood training sets, respectively. As shown in Fig. 5, the selected training points of leaves and wood were marked to red, and were amplified for better visual effects. The overall sampling results have a good performance and meet the expectations. However, tree 2, tree 5 and tree 8 had some mistakenly sampled points, which affected the classification results.

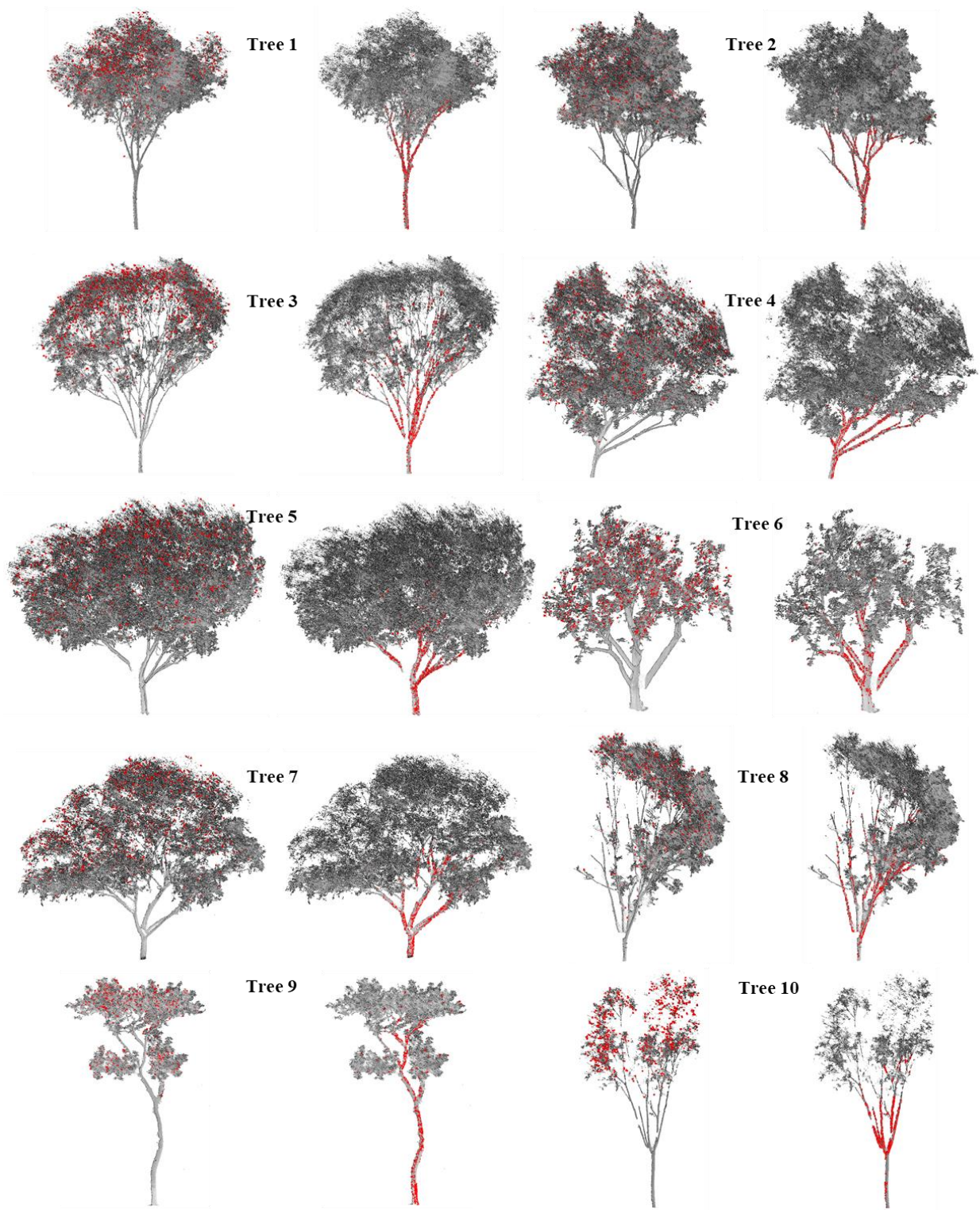

Fig. 5. Automatic selection results of training points.

## 4.2 Classification results

The experiments were carried out by using SVM method. First, leaf and wood training sets selected above were used for training a SVM classifier. Then 10 tree point cloud data were classified into leaf points and wood points by using this classifier. The visual classification results were shown in Fig. 6, in which the leaf points were colored into green and the wood points were colored into brown.

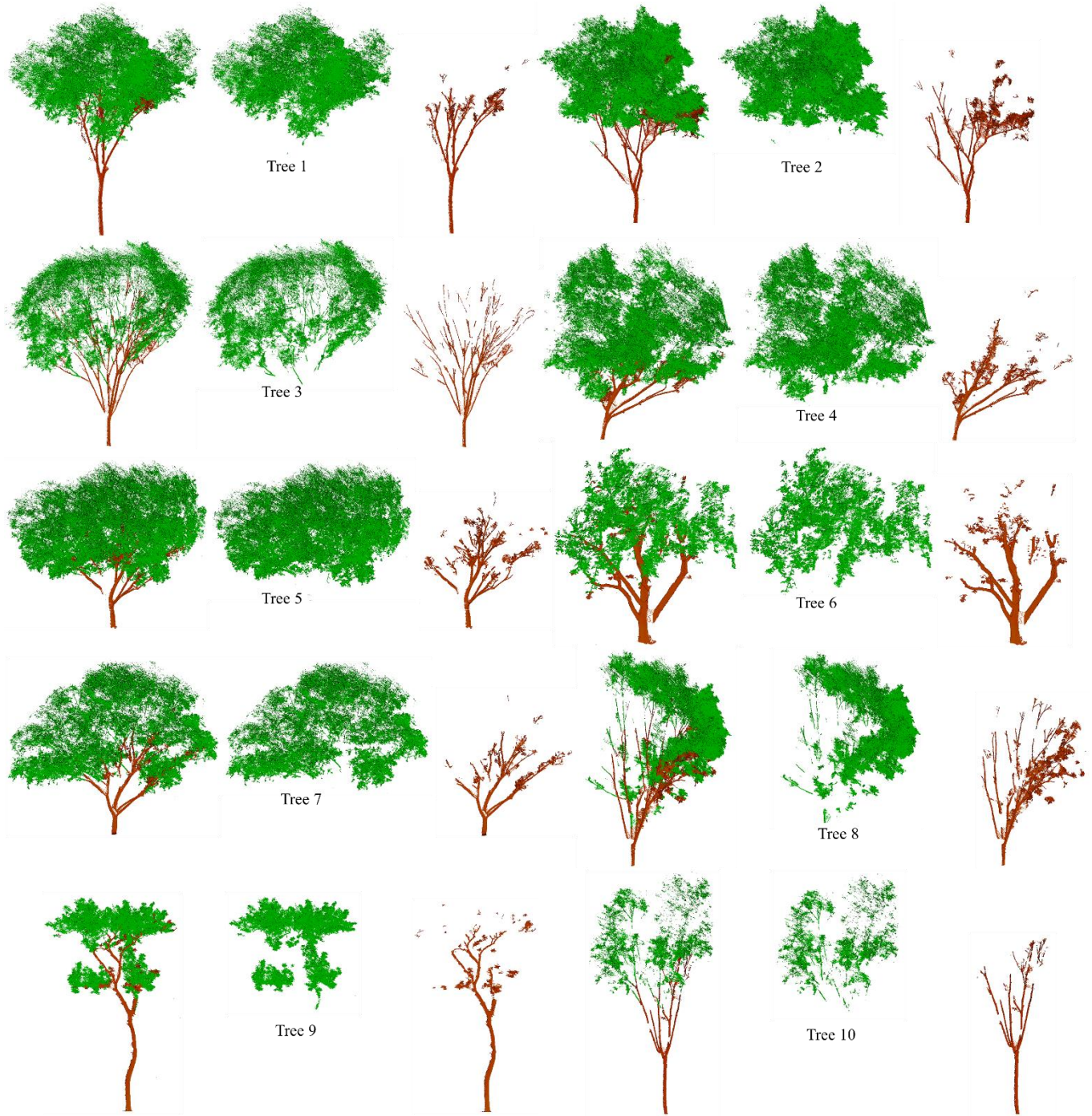

Fig. 6. Classification results by using proposed method.

Then the manual selection method was also used for classification. The numbers of leaf points and wood points in classification results are listed in Table 2, which indicate closer numerical results of the proposed method compared to manual selection method.

| Tree/Numbers | Total numbers of point cloud data | Standard Results | | Automatic method | | Manual selection method | |
|---|---|---|---|---|---|---|---|
| | | Leaf points | Wood points | Leaf points | Wood points | Leaf points | Wood points |
| Tree 1 | 734167 | 641580 | 92587 | 627154 | 107031 | 579396 | 154771 |
| Tree 2 | 2620269 | 2223977 | 396292 | 2098518 | 521751 | 2167950 | 452319 |
| Tree 3 | 549608 | 394893 | 154715 | 427234 | 122374 | 446388 | 103220 |
| Tree 4 | 1586556 | 1322160 | 264396 | 1318359 | 268197 | 1257064 | 329492 |
| Tree 5 | 2098374 | 1838835 | 259539 | 1793534 | 304840 | 1862315 | 236059 |
| Tree 6 | 901094 | 662389 | 238705 | 592002 | 309092 | 559094 | 342000 |
| Tree 7 | 688479 | 559417 | 129062 | 573578 | 114901 | 625261 | 63218 |
| Tree 8 | 2417153 | 1887996 | 529157 | 1796726 | 620427 | 911344 | 1505809 |
| Tree 9 | 1138482 | 863218 | 275264 | 813588 | 324894 | 866545 | 271937 |
| Tree 10 | 210146 | 101521 | 108625 | 113204 | 96942 | 51872 | 158247 |

Table 2. Points numbers of point cloud data and three results.

Besides, the correct classification rates, kappa coefficients and improved accuracy of these two methods were computed by comparing with the standard results for evaluation (as listed in Table 3). As shown in Fig. 6, tree 1, tree 3, tree 8 and tree 10 have neater and clearer classification results than others, since tree 2, tree 8 and tree 9 have some mistakenly classified blocks of results. Tree 8 and tree 10 have higher improvement rates which indicate these two trees have significant improvements of classification effect by using proposed method. Besides, tree 1 and tree 3-7 have mild increases in kappa coefficients. At the same time, tree 2 and tree 9 have little rises in kappa coefficient, because of the mistakenly classified blocks and the potential limitations of this method.

| Tree/Accuracy | Automatic method | | Manual selection method | | Improved accuracy | |
|---|---|---|---|---|---|---|
| | Correct classification rate | Kappa coefficient | Correct classification rate | Kappa coefficient | Correct classification rate | Kappa coefficient |
| Tree 1 | 0.9590 | 0.8258 | 0.8801 | 0.5775 | 7.89% | +0.2483 |
| Tree 2 | 0.9240 | 0.7381 | 0.9283 | 0.7359 | -0.43% | +0.0022 |
| Tree 3 | 0.9243 | 0.8003 | 0.8508 | 0.5897 | 7.35% | +0.2106 |
| Tree 4 | 0.9333 | 0.7612 | 0.8575 | 0.5329 | 7.58% | +0.2283 |
| Tree 5 | 0.9458 | 0.7675 | 0.9163 | 0.5981 | 2.95% | +0.1694 |
| Tree 6 | 0.8961 | 0.7562 | 0.8354 | 0.6287 | 6.07% | +0.1275 |
| Tree 7 | 0.9392 | 0.7917 | 0.8938 | 0.5665 | 4.54% | +0.2252 |
| Tree 8 | 0.9103 | 0.7532 | 0.5512 | 0.2114 | 35.91% | +0.5418 |
| Tree 9 | 0.9354 | 0.8340 | 0.9257 | 0.7965 | 0.97% | +0.0375 |
| Tree 10 | 0.9376 | 0.8755 | 0.7545 | 0.5004 | 18.31% | +0.3754 |
| **Mean** | **0.9305** | **0.7904** | **0.8394** | **0.5738** | **9.11%** | **0.2166** |

Table 3. Kappa coefficients of two methods.

## 5 Discussion

The accuracy evaluation analyzed in experiments showed a good performance of proposed method. As listed in Table 3, the overall correct classification rates of proposed method are better than the manual selection method with 0.9305 compared to 0.8394 in average. Among them, the correct classification rate of tree 2 is slightly lower than the manual selection method but also maintains the same level. In addition, the proposed method greatly improved the kappa coefficients according to the results. Although manual selection method can classify some trees with simple and distinct structures correctly, the proposed method can significantly improve the accuracy of trees with different shapes, which means the proposed method has better universality.

However, there are also some limitations of the proposed method. It relies on the density characteristics of leaves and woods. The sample selection process will be hindered by high-density leaves. For instance, some neighbor points located in high-density leaves may be fitted to a false plane, which will be mistakenly selected as wood sample points, then decrease the classification accuracy. Besides, it may be not applicable in some broadleaved trees because the leaves are shaped like planes themselves.

Some related research publications have also reported their accuracy. Ferrara et al. (2018) classified the point cloud data of 7 cork oak trees and the kappa coefficients were from 0.75 to 0.88. Tao et al. (2015) proposed a method for wood-leaf separation with the kappa coefficients from 0.71 to 0.89. Yun et al. (2016) classified the trees point cloud data of different species by using SVM algorithm, and the overall accuracy ranged from 0.8913 to 0.9349. Zhu et al. (2018) used RF algorithm to classify trees point cloud data and got an average overall accuracy of 0.844 and an average kappa coefficient of 0.75.

Compared with above studies, the proposed method had a similar performance with an automatic process. The correct classification rates of 10 trees range from 0.8961 to 0.9590 and the kappa coefficients of them range from 0.7381 to 0.8755. The means of correct classification rates and kappa coefficients are 0.9305 and 0.7904, respectively.

Obviously, the proposed method can achieve a good classification result without requiring artificial interventions, which may introduce the subjective influence and man-made interference. The automation of the proposed method improves the facility of algorithm. Furthermore, the time cost of feature computation could be reduced by code optimization.

6 Conclusion

The leaf-wood classification of trees plays an important role in forestry and related fields. The proposed automatic classification method can classify tree point cloud into leaf points and wood points efficiently and accurately. Although broad leaf trees or trees with high-density leaves may affect the accuracy of proposed method, it is a feasible and applicable solution for leaf-wood classification. And more future work will be done to improve the efficiency of the method.

## Acknowledgments

This work was supported by the Fundamental Research Funds for the Central Universities (No. 2015ZCQ-LY-02); the State Scholarship Fund from China Scholarship Council (CSC No. 201806515050).